# A Method for Training-free Person Image Picture Generation

Tianyu Chen[1, *]

[1]Faculty of Engineering, University of Bristol, Bristol, The UK

*Corresponding author: uv21031@bristol.ac.uk

**Abstract.** The current state-of-the-art Diffusion model has demonstrated excellent results in generating images. However, the images are monotonous and are mostly the result of the distribution of images of people in the training set, making it challenging to generate multiple images for a fixed number of individuals. This problem can often only be solved by fine-tuning the training of the model. This means that each individual/animated character image must be trained if it is to be drawn, and the hardware and cost of this training is often beyond the reach of the average user, who accounts for the largest number of people. To solve this problem, the Character Image Feature Encoder model proposed in this paper enables the user to use the process by simply providing a picture of the character to make the image of the character in the generated image match the expectation. In addition, various details can be adjusted during the process using prompts. Unlike traditional Image-to-Image models, the Character Image Feature Encoder extracts only the relevant image features, rather than information about the model's composition or movements. In addition, the Character Image Feature Encoder can be adapted to different models after training. The proposed model can be conveniently incorporated into the Stable Diffusion generation process without modifying the model's ontology or used in combination with Stable Diffusion as a joint model.

## 1. Introduction

Image generation models have been gaining a lot of attention in the Artificial Intelligence (AI) field in recent years [1]. In the latter half of 2022, discussions and information regarding these models have been prevalent. In the past, image generation models often required large amounts of video memory. It made it impossible for the people who only had gaming graphics cards, to practice them themselves. The model that is going to break the deadlock in 2022 is Stable Diffusion, a diffusion generation model that has recently shown excellent performance in the field of AI painting. One of the dominant AI painting code frameworks is Stable Diffusion, which is based on an implementation of the Latent Diffusion Model [2].

At this stage, the Latent Diffusion Model is a Diffusion Model that uses various conditions to guide the model generation process (Fig. 1 is a schematic representation of the structure). In the Stable Diffusion implementation, the text is first vectorized by tokenize and subsequently encoded into an encoder hidden states vector by the Contrastive Language-Image Pre-training (CLIP) model [3]. Then, the condition is fed into a part of Stable Diffusion, a UNet [4], along with a generated random seed and a noisy image, in which various Resnet and Attention blocks are used [5]. The condition guides the generation of the final graphics in this process. The image from the UNet is still in Latent space and needs to be decoded by a Variational autoencoder (VAE) to become the final image observed in the AI painting [6]. Due to the VAE, the video memory of the UNet during the processing of the data is thus reduced. This reduces the memory usage of Stable Diffusion to a size that can be run on a normal gaming graphics card. Some of open-source developers have further optimized Stable Diffusion's attention code to run on smaller graphics cards.

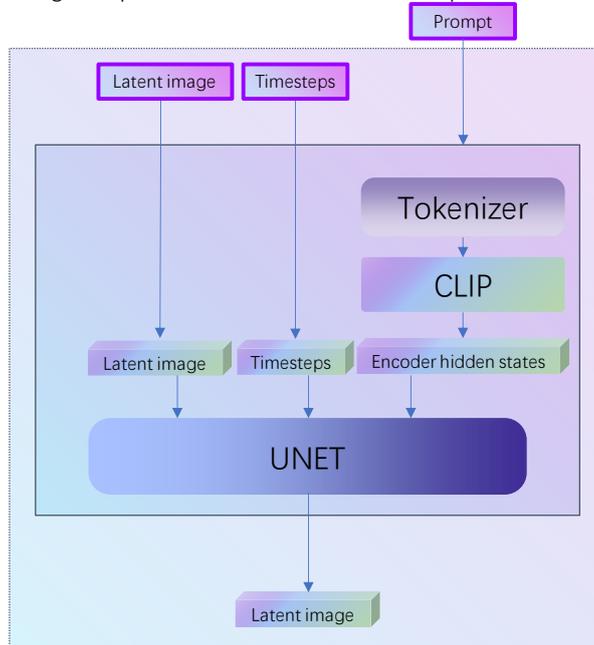

**Fig. 1** The structure of the Latent Diffusion Model.

This is how the entire Stable Diffusion source code shows the process. Such a model shows excellent performance in the practical generation of image results. The fact that Stable Diffusion can run on normal gaming graphics cards has also led to an exponential increase in the number of users and more positive feedback for Stable Diffusion. For example, open-source tools have been developed to make it easier to use Stable Diffusion. However, even though the graphic memory cost of running Stable Diffusion has been reduced to a very low level, there are still some requirements that cannot be met by Stable Diffusion currently. For example, drawing multiple pictures of the same person requires that the images and appearances of the people in the resulting drawings are highly similar. In practice, this can only be achieved by specialized fine-tuning training. Fine-tuning training requires much more equipment and skills than running Stable Diffusion directly with a facilitated open-source tool.

The aim of this paper is to address the problem that drawing a single character during AI image generation requires specialized training to achieve. It enables users to use AI image generation tools to generate various images of the same person without training when drawing a single person, and retaining the original image, appearance.

The training process for Stable Diffusion is similar to the one shown in Fig. 1. Only the VAE-encoded latent image is put into UNet with other parameters after Gaussian noise has been added. The output noise is then calculated as the loss of the noise added to the latent image. Back propagation is then performed to optimize the model weights.

The training process described above fine-tunes both UNet and CLIP, so theoretically it is possible to train a Stable Diffusion model with a high degree of tag freedom, provided that sufficient data is provided. Theoretically, a Stable Diffusion model can be trained with a high degree of tag freedom if sufficient data is provided, and the description is accurate enough. Unfortunately, the Latent Diffusion

Model operates globally on the image, rather than in chunks. Although different drawing distributions are achieved under the guidance of condition, it is still difficult to achieve the same generative results as human will. This means that it is not possible to use pure prompt to infinitely approximate a real individual person or a character from an anime/game.

The new model proposed in this article is therefore partly designed to solve the problem that a character image needs to be trained specifically for this purpose if it is to be drawn well. This model is called **Character Image Feature Encoder** (which can be abbreviated as CIFE or Character Encoder). The purpose of this model is to change the Stable Diffusion so that it can be used by the average user. It is possible to draw a given character without training, but only a portrait of the character. and some prompts describing the result the user wants. Very often some characters do not have a lot of training material, and the time and equipment costs to train the model are not bad. So, this model can be a good change from that.

## 2. Method

### 2.1 Technical overview

To achieve training-free image generation of a given character with only a picture of the character, it is necessary to input a batch of information to the model and have the model draw the same prompt with these hidden conditions. The most effective approach involves creating a single input that is identical to the CLIP encoding result. Since the Stable Diffusion implementation of the model does not limit the length of the CLIP encoding result. The classical length at this stage is generally (77,768), but it can be longer. This means that the condition vector for that value input can have more than just the CLIP result, or that CLIP can have a parallel encoder present. Only CLIP encodes the vectorized text as a hidden vector. The encoder, which works in parallel with the CLIP encoder, is encoding the visual information of the character as a hidden vector.

### 2.2 Model Structure

According to the tone and strategy set out above, in order to achieve the encoding of images as Hidden stats to be fed into UNet together with the CLIP encoding results, a network structure for feature extraction of image information is firstly required, followed by some encoded neural network layers for further character image feature extraction of the extracted image features, and finally as encoding results to be directed to UNet together with CLIP UNet to generate the images. The model will therefore be divided into two main parts, the image feature extraction structure, and the deep feature coding structure. A schematic of the structure of the model can be seen in Fig. 3 below.

### 2.3 Image Feature extraction

The image feature extraction process is well established in the industry [7-9], but it is a combination of various CNN layer networks, such as VGG networks or ResNet networks [5, 10], which are excellent classification models, but these models can also be used by the Character Encoder. layer can directly use the feature layer of these networks as the feature extraction layer, which has the advantage of using existing pre-trained weights for migration learning, in addition to being a tried and tested structure for image feature extraction networks. Although there is a difference between the pre-training weights for the feature extraction network of the classification network and the weights required for character encoding characters, there are always similarities and differences, which means that using migration learning can be less costly in terms of time.

### 2.4 Deep Character Image Encoder

The encoding process of the deep character image is the most important part of this model. Deep character encoding is just a matter of feeding the encoding results of the feature layer into a neural network of linear layers for processing. Only an appropriate number of neurons and network depth are required to encode the image features to the persona vector.

### 2.5 Training of Character Image Feature Encoder

The aforementioned structures form a model with full image feature extraction + character encoding capabilities, forming a complete Character Encoder model. For the training of the model, the best way is to adapt it directly to the existing Stable Diffusion model. Moreover, because it is a homogeneous model of the CLIP model, the Character Encoder can be directly adapted to Stable Diffusion models with different weights. The freedom of substitution does not affect the original weights of the model, which is also fundamentally different from fine-tuning training. As a direct adaptation of the training method to existing Stable Diffusion models, it is only necessary to plug the model directly into the Stable Diffusion model training process described in the previous section. The CLIP output is then stitched together with the Character Encoder output and fed into UNet (as shown in Fig. 3).

## 3. Result and Discussion

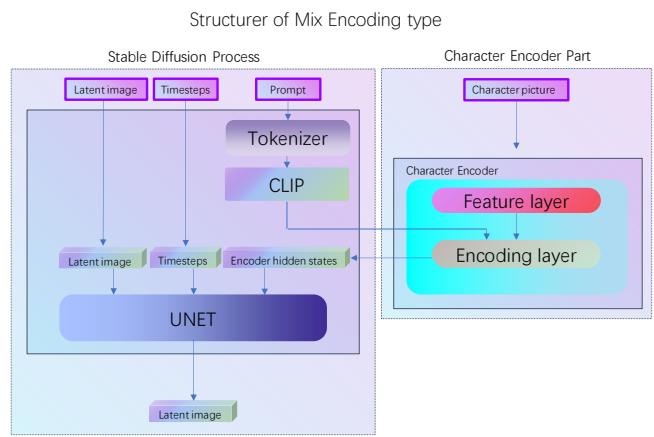

**Fig. 2** The structure of mix encoding type.

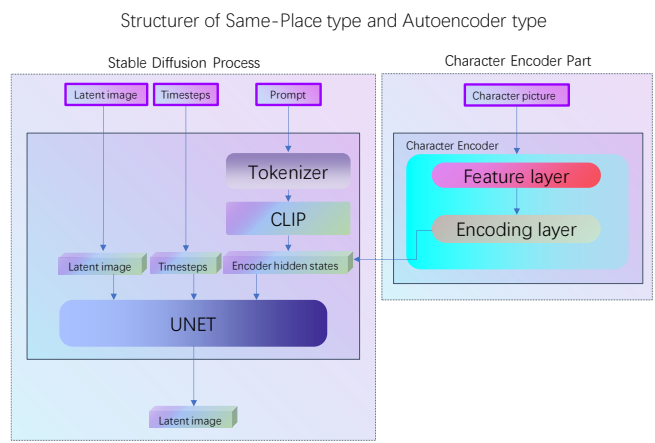

**Fig. 3** The structure of same-place type and autoencoder type.

The experimental models constructed according to the above theoretical guidelines can be divided into three main categories, namely Mix Encoder type, Same-place type and Autoencoder type, both of which have the same image feature extraction part and use the special diagnostic extraction layer of the VGG model directly. However, they differ in the part of character encoding. All three models use the feature extraction layer to extract features from the image when combined with the Stable Diffusion model. The difference between the three models lies in the encoding part. As shown in Fig. 2, the Mix Encoder type encodes the CLIP results and the Character Encoder results after concatenation, while the Same-place type is shown in Fig. 3. The Autoencoder type is special. It has the same model structure as the Same-place type when combined with Stable Diffusion (Fig. 3), but the pre-training process (Fig. 4) can be performed separately from the Stable Diffusion model set.

### 3.1 Training environment

#### 3.1.1 Dataset

The dataset for training includes 18 characters (All from a game that called "Arknights"). There are 2-4 pictures for character appearance image $C$ and 50-70 images that is fanart $F$. The character appearance pictures will be input into character encoder for extract character image feature, and fanart images is preparing for Stable Diffusion training process. Therefore, the total dataset is: $\sum_{i}^{18} C_i * F_i$.

#### 3.1.2 Other environments

The basic model of training is Anything-V3.0, the training hardware platform is based on 3090, the version of PyTorch is 1.12.1.

### 3.2 Mix Encoder type

#### 3.2.1 Result

The training results of the Mix Encoder type model were not satisfactory, with the training results outputting images that were incomprehensible. It same to in the case of the training failure caused by some misuse of the Textual Inversion training. In addition, the output of Stable Diffusion model that generates process guide by character encoder encoding result show us completely randomness. Which means the result was not get influence by character encoding information.

#### 3.2.2 Discussion

Although the Mix Encoder type model has complete fail in the above results, the reason is the training did not have enough training data and training steps. Because the dataset of CLIP has nearly 400 million images. However, the dataset of this experiment only has thousands of data after the character image data is combined with fanart image. Therefore, the reason for Mix Encoder type model failure may not be the technical root of Mix Encoder type is fault, it is the dataset not enough.

### 3.3 Same-place type

#### 3.3.1 Training details

The training process of the model blocks the weights of the UNet and clip models, and the optimizer only adjusts the weights of the character encoder model.

#### 3.3.2 Result

This is the best performing model of the three types of models. The training results of the Same-place model show good quality of generated images. In addition, the images of the character persons are also well represented in the generated images. In the test of character image encoding ability. The test chooses a picture that was not in the training set, and the character image in the picture was also not in the training set. The result of the test was that the generated picture clearly brings that the character image encoded in the picture. In addition, after changing the main model, the generated picture also carried the encoded character image.

Fig. 4 Image 1 in the following images is the image of the character used to input the Character Encoder. Fig. 4 Image 2 is result of the character drawn by the Stable Diffusion model Anythin-V3.0 guided by the encoding results of the Character Encoder. Fig. 4 Image 3 is a picture of the character drawn from the AbyssOrangeMix2 model guided by the encoding results. Fig. 4 Image 2 is a picture of basil mix model (real-life model) guided by the b-encoding results. The last image, on the other hand, is used as a control group. The images generated with the exact same parameters on Anything-V3.0 **without Character Encoder enabled**. These images are drawn with the following parameters: Seed: 7313187166, sampler: DDIM, sample steps: 30. Where the only variables for pictures 2-4 are different master models only.

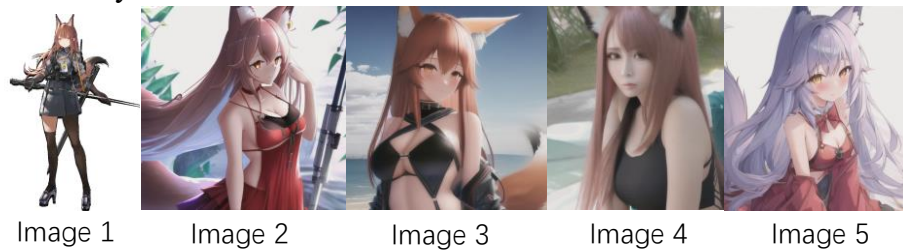

Image 1　　Image 2　　Image 3　　Image 4　　Image 5

**Fig. 4** The sample images used in this study.

### 3.3.3 Discussion

The results of the experiments show that the image of the person is still reflected in the generated image after replacing the model as described above (Fig. 4 Image 2-Image 4). In addition, the Image is the result of Anything-V3.0 model generation without Character Encoder, this can prove the Character Encoder has the influence on the process of generation. The results of the training of the Same-place model demonstrate the viability of this route of the Character Encoder and prove that it is possible to implement this technique. The result of the encoding of the character encoder carries the data of the character's image. Therefore, Character Encoder achieves its intended purpose. Furthermore, CLIP and UNet were turned off during the training process. This avoids the character image being trained by CLIP or UNet on the combination of prompts, because it will evade the optimization of the character encoder. This is because UNet is well suited to fine tuning with small data sets. The role encoder, on the other hand, is trained with completely unadjusted weights except for migration learning from the VGG pre-trained model. The optimizer will choose to fine-tune UNet and CLIP rather than train the role encoder to get there faster. And, according to the result that previous description. Turning off UNet and CLIP training has the added benefit that the role encoders trained in this way can be applied to a different master model, rather than being restricted to the same master model. This is also well reflected in the training results.

## 3.4 Autoencoder type

### 3.4.1 Special model structures

The Autoencoder model is used in the same way as the Same-place model. However, there are two stages in the training process. The first stage is the training of the Autoencoder, where the encoder part is the same as the homunculus model, but the decoder part is not the same as the decoder usually used in industry. This is because the decoder part introduces CLIP encoder hidden states to differentiate between the different content of the same Same-place model, to guide the encoder to better extract the character features. In the second stage, the trained encoder and Stable Diffusion are trained together to fine-tune the model.

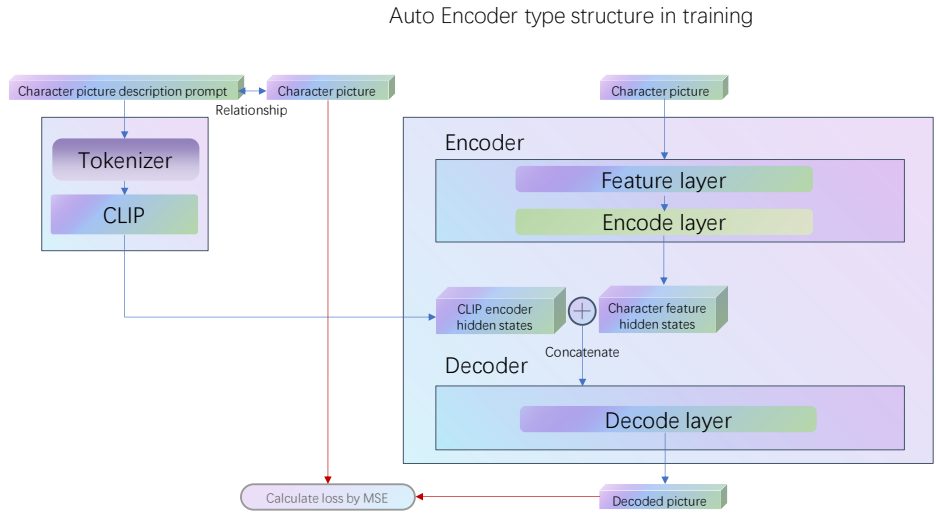

**Fig. 4** The auto encoder type structure in training.

### 3.4.2 Result

The Autoencoder type training yields results that, while superior to mixed encoding type training, fall short of the quality of Same-place type generated results. Nonetheless, the Autoencoder type still generates outputs that distinctly showcase character features, albeit of inferior quality compared to Same-place generated images.

### 3.4.3 Discussion

The reason why the Autoencoder type model is not as effective as the Same-place model but is mentioned anyway is that the training log for the second training phase of the autoencoder is not the same as the training log for the Same-place model. The autoencoder training process begins with the appearance of distinct persona features in the generated samples. This indicates that the first stage of autoencoder training was effective and succeeded in making the encoder output contain the extracted persona features.

## 4. Conclusion

Based on the experimental results, it is evident that the technical approach of the figure encoder is feasible. The same-place conformation model exhibits exceptional character extraction encoding capabilities and can be applied to different master models. However, the success of the Same-place conformation does not mean that the other two conformations fail. As mentioned in the previous discussion, the analysis suggests that this is a problem due to the insufficient amount of data and training steps. It is believed that if the amount of data can be increased, then the other two configurations should be able to perform as well as the Same-place configuration. Therefore, the following directions are possible for the future development of this technique. The first priority is to increase the amount of data and the training time. Secondly, there are two derivative directions. The first is to train the Character Encoder alone, as mentioned above, and only adjust the weights of the Character Encoder when trained alongside the Stable Diffusion model. The other direction is to fine-tune the overall Stable Diffusion model after the Character Encoder has been trained. For the Mix Encoding fine-tune, the Character Encoder, UNet and CLIP can be trained simultaneously. While for the AutoEncoder configuration, it can only train the Character Encoder and UNet.

However, the future direction of this technology will also have a larger application scenario,

which is to train on real-life datasets. In contrast to anime drawings, rea-life dataset should be better. Because the real-life data sets are taken by cameras and the errors are minimal and there are gaps between the drawings of different artists. Theoretically, this would result in a better implement than the animated characters. Perhaps the Character Encoder trained on the live action dataset could in turn be transferred to learn the application scenario of animated character drawing. This would in turn overtake a model trained on a purely animated character dataset.